\documentclass[11pt]{article}

\usepackage[final]{acl}

\usepackage{times}
\usepackage{latexsym}

\usepackage[T1]{fontenc}

\usepackage[utf8]{inputenc}

\usepackage{microtype}

\usepackage{inconsolata}

\usepackage{graphicx}

\usepackage{multirow}

\usepackage{makecell}
\usepackage{rotating}

\usepackage{amsmath}

\title{"Înțelegi românește?" A Recipe for Romanian Vision-Language Models}

\author{
 \textbf{Mihai Masala\textsuperscript{1}},
 \textbf{Marius Leordeanu\textsuperscript{1,2}},
 \textbf{Mihai Dascalu\textsuperscript{1}},
 \textbf{Traian Rebedea\textsuperscript{1}}
\\
 \textsuperscript{1}National University of Science and Technology POLITEHNICA Bucharest, Romania,
}

\begin{document}
\maketitle
\begin{abstract}
Vision-Language Models (VLMs) largely follow the text-only LLM trajectory, excelling on English benchmarks but sharply degrading on low-resource languages, where neither large-scale image-text corpora nor culturally grounded evaluations exist. We present a systematic study of building a language-specific VLM for Romanian, covering the full pipeline from data construction to architectural choices. We translate established English VLM training and evaluation corpora into Romanian, applying machine translation to textual annotations and to in-image text, preserving visual grounding while adapting the textual content. Using this data, we train and ablate a series of VLMs to isolate the contribution of (i) vision backbones of varying scale and pretraining, (ii) language backbones from multilingual to Romanian-adapted LLMs, and (iii) OCR-style image-text data. We further curate HoraVQA, a culturally native evaluation set grounded in Romanian everyday scenes. Romanian-adapted VLMs consistently outperform their same-sized counterparts and, across all evaluated benchmarks, even surpass models from the next larger size category. 


\end{abstract}

\section{Introduction}

Recent Vision-Language Models (VLMs) have achieved strong multilingual capabilities, largely driven by advances in multilingual Large Language Models (LLMs) and large-scale multimodal pretraining. However, it remains unclear how effectively these capabilities transfer to grounded multimodal understanding in lower-resource languages~\cite{liu2021visually}. While current systems can often generate fluent responses in many languages, tasks involving OCR, document image understanding, culturally grounded reasoning, and localized visual semantics remain insufficiently studied.

A major challenge is the lack of comprehensive multimodal resources and evaluation frameworks for many languages. Existing datasets often under-represent naturally occurring image-text pairs, domain-specific documents, and culturally relevant visual contexts, making it difficult to systematically assess model behavior beyond general conversational ability. As a result, there has been a growing effort to develop language- or region- specific multimodal benchmarks~\cite{hsieh2026taiwanvqa,cahyawijaya2025crowdsource} to more accurately evaluate model performance across diverse linguistic and cultural contexts. 

In this work, we take a step further by not only studying cross-lingual multimodal evaluation, but also examining the role of language-specific training data, including translated and aligned image–text pairs, in improving VLM performance for low-resource languages. 
We also construct a comprehensive  evaluation suite that contains 19 benchmarks spanning OCR, visual question answering, captioning, and reasoning tasks, and investigate adaptation strategies for multilingual vision-language models. Our experiments show that strong multilingual text capabilities do not consistently translate into robust multimodal performance, particularly in OCR-intensive and culturally grounded settings. We further analyze how data composition and instruction tuning affect low-resource multimodal transfer.

Our contributions can be summarized as follows:

\begin{itemize}
    \item We introduce a comprehensive Romanian multimodal evaluation suite, consisting of 19 benchmarks spanning OCR, visual question answering, captioning, and reasoning, including a novel fully human-annotated culturally grounded test set.
    \item We conduct a systematic study of training strategies for multilingual VLM adaptation, including ablations over text and vision backbones, and analyze the role of OCR and data composition in low-resource multimodal performance. 
    \item We introduce HoraVQA ("\textit{hora}" is a Romanian circle dance that signals community and tradition), a culturally native evaluation set of more than 500 question-answer pairs grounded in Romanian everyday scenes;
    \item We release RoVLM models for Romanian multimodal understanding along with a fully open and reproducible training and evaluation pipeline, including data processing, training recipes, and evaluation protocols.\footnote{\url{www.openllm.ro}}
\end{itemize}

\section{Related Work}

Vision-language models (VLMs) have rapidly evolved from contrastive image-text representation learning approaches such as CLIP~\cite{radford2021learning} into large instruction-tuned multimodal systems capable of visual reasoning, captioning, and dialogue~\cite{gemmateam2025gemma3technicalreport,wang2024qwen2}. Modern VLMs couple a vision encoder to an LLM and are trained with visual instruction tuning, LLaVA~\cite{liu2024improved}, InstructBLIP~\cite{dai2023instructblip}, and recent open families such as Qwen-VL~\cite{wang2024qwen2,bai2025qwen25vltechnicalreport}, InternVL~\cite{chen2024internvl,wang2025internvl3}, and PaliGemma~\cite{beyer2024paligemma}. We adopt this modular recipe as the backbone for Romanian adaptation.

Alongside these advances, recent work has increasingly explored multilingual extensions of VLMs. Multilingual coverage has been pursued either at pretraining scale~\cite{chen2022pali,chen2023pali,chen2023altclip,tschannen2025siglip} or via translated instruction data: M3IT~\cite{li2023m}, PALO~\cite{maaz2024palo}, and Maya~\cite{alam2024maya}. Aya Vision~\cite{dash2025aya} adds synthetic multilingual annotation and cross-modal model merging to limit text-only forgetting, and LRM-LLaVA~\cite{li2025lrm} argues that residual gaps stem from a modality gap between visual inputs and non-English text. \cite{hinck2024llava} further show that LLaVA-style models drift toward English responses when an image is present -- a failure mode our Romanian-only adaptation directly targets.

Several existing works adapt VLMs to a single language: LLaVA-NDiNO for
Italian~\cite{musacchio2024llava}, LaVy for Vietnamese~\cite{tran2024lavy}, Qolda for Kazakh~\cite{arystanbekov2026qolda} and, closest to us ~\cite{dima2025parameter} who LoRA-tune LLaMA/LLaVA/Qwen variants on a translated Romanian Flickr30k. RoVLM differs in scope: a 3.1M-sample mix containing alignment, captioning, instruction VQA, OCR/documents, grounding data, 19-benchmark evaluation, and in-image text translation.

Parallel evaluation probes how VLMs behave across languages and cultures. Early work translated tasks or  multilingual captions and VQA, including MaRVL~\cite{liu2021visually}, Crossmodal-3600~\cite{thapliyal2022crossmodal}, and MaXM~\cite{changpinyo2023maxm}. More recent benchmarks shift toward culturally grounded content: CVQA~\cite{cvqa}, CulturalVQA~\cite{nayak2024benchmarking}, ALM-Bench~\cite{vayani2025all}, EXAMS-V~\cite{das2024exams}, TaiwanVQA~\cite{hsieh2026taiwanvqa}, and SEA-VL~\cite{cahyawijaya2025crowdsource}. The evaluation suites AyaVisionBench~\cite{dash2025aya} and m-WildVision~\cite{dash2025aya} translate prompts over shared images; we complement them with \textbf{HoraVQA}, a Romanian-authored benchmark sourced from Wikidata/Wikimedia with culturally tagged questions.

Prior multilingual VLM work has pursued either broad multilingual coverage through large-scale pretraining or instruction tuning in translated form, or benchmark construction to measure cross-lingual and cross-cultural gaps. In contrast, our work studies the full adaptation pipeline for a single low-resource language. We focus on Romanian and combine translated multimodal supervision, translated in-image text, native Romanian OCR/document data, and a human-authored cultural benchmark. This setting allows us to isolate which components of adaptation matter: the language backbone, the vision backbone, OCR-heavy supervision, and text localization within images. Our findings show that Romanian multimodal performance depends not only on multilingual language modeling, but also on OCR-sensitive data composition and grounded cultural coverage.

\section{Dataset Development}

We collect image–text data from a diverse set of English and Romanian-language sources to enable language-specific multimodal training and evaluation. Our dataset includes naturally occurring image-caption pairs, OCR-rich visual content, instructional data, and document-style tasks covering a broad range of domains and visual formats. In addition to native Romanian data, we incorporate translated image–text pairs to study the impact of language-specific supervision in low-resource multimodal adaptation.

We rely on established, high-quality datasets that are usually used for instrution-tuning and evaluating strong multimodal models. Our  aim is to study the impact of using translated versions of this data for Romanian VLM training. In total, we selected 11 training datasets as follows: LAION~\cite{schuhmann2022laion}, LLaVA-Mix~\cite{liu2024improved}, PixMo~\cite{deitke2025molmo}, Flickr30K~\cite{dima2025parameter}, CoSyn~\cite{yang2025scaling}, and FinePDFs~\cite{kydlicek2025finepdfs}. In the end, we were left with 3.17M training samples, with datasets grouped per task in Table~\ref{tab:train_datasets}. We observe substantial variation in text complexity and visual structure across domains (see details in Appendix~\ref{sec:appendix-training-data}), motivating the need for diverse multimodal evaluation beyond standard captioning settings.

\begin{table}[hbt]
  \centering
  \begin{tabular}{llr}
    \hline
    \textbf{Task} & \textbf{Dataset} & \textbf{\# Samples} \\
    \hline
    Alignment
        & LAION            & 552k \\
    \hline
    \multirow{2}{*}{Captioning}
        & PixMo-Cap       & 601k \\
        & Flickr30K-Cap    & 25k \\
    \hline
    \multirow{4}{*}{\shortstack[l]{General VQA \&\\Instruction}}
        & LLaVA-Mix       & 618k \\
        & PixMo-AA        & 131k \\
        & PixMo-CapQa   & 205k \\
        & Flickr30K-Qa     & 25k \\
    \hline
    \multirow{2}{*}{\shortstack[l]{OCR \&\\Documents}}
        & CoSyn            & 422k \\
        & FinePDFs         & 375k \\
    \hline
    \multirow{2}{*}{Grounding}
        & PixMo-Count     & 34k \\
        & PixMo-Points    & 185k \\
    \hline
    \textbf{Total} & & \textbf{3{,}1M} \\
    \hline
  \end{tabular}
  \caption{\label{tab:train_datasets}
    Training data grouped by the capability each source contributes. All images, instructions, and outputs are in Romanian, obtained by translating the original English datasets (where applicable).}
\end{table}

\subsection{Translation and Adaptation of Existing Datasets}
\label{sec:translation}

As Romanian multimodal supervision remains limited compared to high-resource languages (FinePDFs is the only dataset that contains native Romanian data), we resort to machine translation. Compared to previous approaches~\cite{musacchio2024llava,tran2024lavy,maaz2024palo}, in this work, we also investigate \textit{"translating"} and adjusting the visual content, the images, not just the texts. This enables us to analyze the extent to which translated supervision can support multimodal transfer in low-resource settings. For OCR-sensitive tasks, we use native language, real-world data in the form of FinePDFs, together with synthetically generated charts and tables adapted from CoSYN.

For deciding what model to use for machine translation, we perform a small-scale experiment: we evaluate five translation models on a subset of MMMU~\cite{yue2024mmmu} data with \textit{gemini-2.5-flash} and \textit{claude-3-7-sonnet-20250219} as judges. We select both open source models -- LLMic~\cite{buadoiu2025llmic}, Seed-X-PPO~\cite{cheng2025seedxbuildingstrongmultilingual}, and closed models -- GPT-4o-mini\footnote{gpt-4o-mini-2024-07-18}, GPT-4.1-mini\footnote{gpt-4.1-mini-2025-04-14} and DeepL\footnote{www.deepl.com/en/translator, last access 19th May 2026}. Based on the results in Table~\ref{tab:translation_score}, we select GPT-4.1-mini for translating benchmarks (cheaper and just slightly lower performance compared to DeepL) and Seed-X-PPO for textual translating training data. 

\begin{table}[hbt]
  \centering
  \begin{tabular}{lcc|r}
    \hline
    \textbf{Model} & \textbf{Gemini} & \textbf{Claude} & \textbf{Avg}\\
    \hline
    LLMic & 8.08 & 7.96 & 8.02 \\
    Seed-X & 8.50 & 8.39 & 8.45 \\
    4o mini & 8.61 & 8.83 & 8.72 \\
    4.1 mini & 8.85 & 8.83 & 8.84 \\
    DeepL & 8.81 & 8.99 & 8.90 \\
    \hline
  \end{tabular}
  \caption{\label{tab:translation_score}
    Translation performance of different models.}
\end{table}

Translating visual input is performed in three steps: a) extracting text from the image; b) translating the text; c) replacing the original text with the translated one, maintaining as much as possible its placement, font, and size. The entire process is performed using custom adaptations of open-source toolkits~\footnote{https://github.com/boysugi20/python-image-translator, last accessed 19th May 2026}. Language translations were validated by human annotators using two quality filters: whenever there was a significant (1.5x) length difference between the original and translated text, and an additional random sampling over all translations. For examples and more details about the translation process, see Appendix~\ref{sec:appendix-translation-process}.

Two notable exceptions stand out: FinePDFs and CoSyn. The former contains native Romanian data (in the form of PDFs), data that we use for pure OCR. The CoSyn dataset contains synthetically generated charts and tables, while also including the Python source code. In this case, rather than translating the image itself—which is generated programmatically—we translate the visualization-related strings, then regenerate the output to ensure high-quality charts.

\subsection{HoraVQA: A Native Romanian Cultural VLM Benchmark}
\label{sec:HoraVQA}

\textbf{Motivation.}
Existing multilingual VLM benchmarks (e.g., MMBench~\cite{liu2024mmbench}, MMMU~\cite{yue2024mmmu}) are overwhelmingly English-centric, with non-English coverage produced through post-hoc machine translation of the textual fields and, in our case, of in-image text. This process inherits the cultural distribution and potential biases of the source corpus and tests visual-linguistic recognition in a culturally generic setting. Even multilingual collections that include Romanian, such as AyaVisionBench~\cite{dash2025aya} and m-WildVision~\cite{dash2025aya}, contain Romanian items but with images shared across all languages and therefore unrelated to the Romanian cultural context. To our knowledge, no existing benchmark probes whether a VLM \emph{understands Romanian-specific visual culture}---monuments, cuisine, folk traditions, visual heritage, or national symbols. HoraVQA is intended to close this gap and to provide an evaluation for day-to-day user queries embodied in images related to the Romanian cultural and historical landscape. 

\textbf{Image sourcing.}
We seeded a list of $503$ Romanian cultural concepts from Wikidata and Wikipedia spanning themes such as 
heritage sites, traditional cuisine, folk customs, national symbols, or recent history iconography. For each
concept, we queried Wikimedia Commons and retained openly-licensed images that visually depict the concept unambiguously. After a manual triage pass that removed duplicates, mislabeled images, and selected items representative of Romanian culture, the final image pool contains $438$ images, manually split into $9$ mutually exclusive categories: landmarks, food \& drinks, customs, daily life, transport, arts, sports, people, and recent history.

\paragraph{Question-answer pairs.}
Questions were authored \emph{natively in Romanian} by 7 volunteer annotators using a custom web tool. Each annotator was instructed to write, for each image of their choosing (while allowing them to skip any image), at least one question either as multiple-choice (four options, exactly one correct) or open-ended free-form answer. This annotation pipeline yielded $580$ question-answer pairs on $232$ unique images: $394$ MCQ and $186$ open-ended, with an average of $2.5$ questions per image (min $1$, max $6$). A question may share an image with Romanian cultural content, yet itself requires no such knowledge to answer (e.g.,\ ``How many people appear in this image?''). We therefore ask users to tag every question they created with a binary \textit{is\_cultural} flag probing whether a correct answer requires knowledge of Romanian culture, history, geography, or language beyond generic visual perception. More details about the annotation process are presented in Appendix~\ref{sec:appendix-annotation-process}.

\paragraph{Dataset statistics.}
Table~\ref{tab:HoraVQA-stats} summarizes the dataset statistics. The benchmark follows a long-tail category distribution without collapsing onto a few dominant classes: the largest category (\textit{landmarks}) accounts for $27.9\%$ of questions, while no category falls below $4\%$. Cultural-question density varies
across categories, with \textit{recent history} ($68\%$), \textit{customs} ($65\%$), \textit{people} ($65\%$) and \textit{arts} ($62\%$) being the most knowledge-intensive, and
\textit{daily\_life} ($46\%$) and \textit{transport} ($48\%$).

\begin{figure}[hbt]
  \includegraphics[width=\columnwidth]{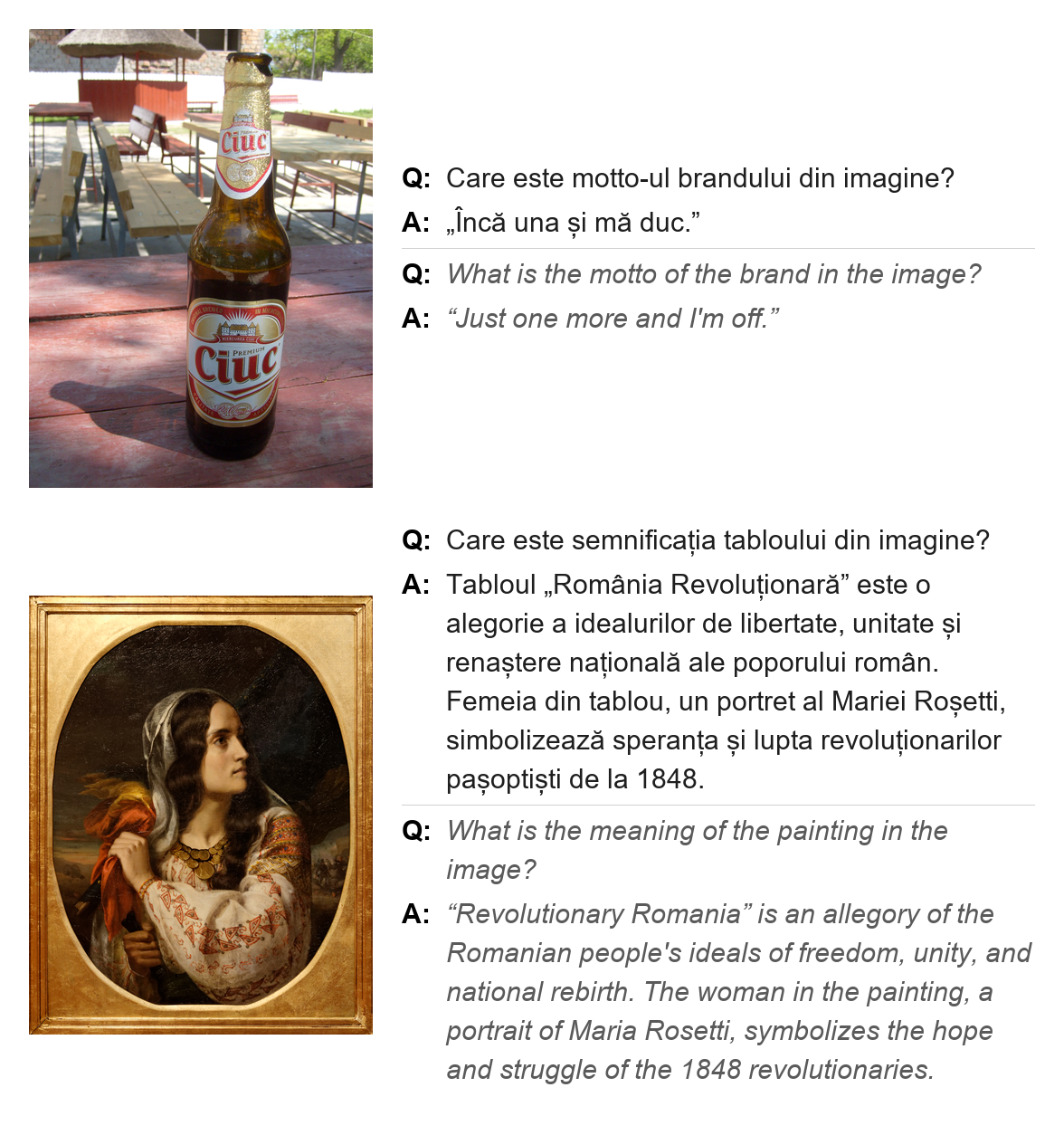}
  \caption{QA examples from the HoraVQA benchmark, shown in their original form (top) and translated into English (bottom). The first example refers to a well-known commercial associated with the brand depicted in the image, while the second concerns a famous painting portraying the ideals of Romanian society around 1850.}
  \label{fig:HoraVQA_example.png}
\end{figure}

The two examples in Figure~\ref{fig:HoraVQA_example.png} illustrate the breadth of cultural knowledge HoraVQA targets: the first probes recognition of a contemporary commercial brand and its associated advertising, while the second requires identifying a 19th-century painting and reasoning about the social ideals it depicts. Together, they span pop-culture familiarity, historical art recognition, and period-specific social context.


\begin{table}
  \centering
  \small
  \setlength{\tabcolsep}{4pt}
  \begin{tabular}{lrrcc}
    \hline
    \textbf{Category} & \textbf{\#Img} & \textbf{\#Q} & \textbf{MCQ\,/\,Open} & \textbf{Cult\,/\,N-cult} \\
    \hline
    landmarks & 63 & 162 & 109\,/\,53 & 86\,/\,76 \\
    food \& drinks & 44 & 107 & 71\,/\,36 & 62\,/\,45 \\
    customs & 34 & 81 & 56\,/\,25 & 53\,/\,28 \\
    daily life & 23 & 56 & 43\,/\,13 & 26\,/\,30 \\
    transport & 21 & 46 & 32\,/\,14 & 22\,/\,24 \\
    arts & 15 & 39 & 20\,/\,19 & 24\,/\,15 \\
    sports & 11 & 33 & 24\,/\,9 & 19\,/\,14 \\
    people & 12 & 31 & 24\,/\,7 & 20\,/\,11 \\
    recent history & 9 & 25 & 15\,/\,10 & 17\,/\,8 \\
    \hline
    \textbf{Total} & \textbf{232} & \textbf{580} & \textbf{394\,/\,186} & \textbf{329\,/\,251} \\
    \hline
  \end{tabular}
  \caption{\label{tab:HoraVQA-stats}
    HoraVQA composition. For each image-level category we report the number of distinct images, total questions, the multiple-choice / open-ended split, and the cultural / non-cultural split.}
\end{table}



\section{RoVLM Models}

For training the RoVLM models, our aim was to build upon existing multilingual vision-language model architectures composed of a vision encoder, a multilingual language model backbone, and a multimodal projection module. Rather than introducing a new architecture, we systematically study the adaptation of modern multilingual VLMs to Romanian multimodal understanding tasks. We explore multiple combinations of vision and language backbones and training dataset mixes to evaluate their impact across diverse downstream tasks.

To adapt multilingual VLMs to Romanian, we employ instruction tuning on Romanian-centric multimodal data. Our training setup incorporates translated image–text and responses pairs alongside naturally occurring Romanian supervision, enabling controlled analysis of language-specific data effects in low-resource settings. We further investigate the role of OCR-rich samples and task diversity in improving grounded multimodal understanding.

For most of the experiments, including the ablation studies mentioned before, we resort to LLaVA-NeXT~\cite{liu2024improved} architecture for its simplicity and modularity.  Besides LLaVA-NeXT, we fine-tune models from three additional VLM families under the same recipe: Qwen2-VL~\cite{wang2024qwen2}, Qwen2.5-VL~\cite{bai2025qwen25vltechnicalreport}, and Gemma3~\cite{gemmateam2025gemma3technicalreport}. This spans two vision encoder families (CLIP-style and SigLIP), different connector designs, and three distinct language backbones, letting us test whether our findings generalize across architectures rather than being tied to a single design.

\subsection{Training Setup}
For all experiments, we used the same training setup: code-base, data mix (see Table~\ref{tab:train_datasets}, and hyperparameters. Following recent multimodal models~\cite{wang2024qwen2,gemmateam2025gemma3technicalreport}, we omit a dedicated alignment phase, directly training with the vision backbone frozen, while the adapter and language backbone are trainable. We train for one epoch with the AdamW optimizer and a cosine learning rate scheduler with a peak of $1.0\times10^{-5}$ and a minimum learning rate $10\times$ lower, preceded by a warm-up phase covering $2.5\%$ of total steps. The effective batch size is 64, with a maximum sequence length of 8192 tokens. Full hyperparameters are listed in Appendix~\ref{sec:appendix-hparams}.

\subsection{Evaluation Suite}
We evaluate on 19 benchmarks spanning six capability groups (Table~\ref{eval-benchmarks}), totalling 59{,}840 samples. \textit{General understanding}, with MMBench~\cite{liu2024mmbench}, MMStar~\cite{chen2024we}, SeedBench2~\cite{li2024seed} and \textit{Knowledge \& reasoning} with MMMU~\cite{yue2024mmmu} and MME~\cite{fu2026mme} probe broad multimodal competence. \textit{Cultural} coverage combines CVQA~\cite{cvqa} and ALM-Bench~\cite{vayani2025all} with the native Romanian RoMemes~\cite{puaics2024romemes} and HoraVQA; CVQA and RoMemes are treated as closed-form classification, while ALM-Bench and HoraVQA — both of which contain open-ended items — are scored by an LLM judge. \textit{Generation \& open-ended} benchmarks include Flickr30k-Caption~\cite{dima2025parameter}, scored with BLEU, ROUGE, and BERTScore, and Flickr30k-QA, LLaVA-Wild~\cite{liu2024improved}, AyaVisionBench~\cite{dash2025aya}, and m-WildVision~\cite{dash2025aya}, all scored by an LLM judge. \textit{OCR \& Document} test text extraction: CoSYN~\cite{yang2025scaling} is judge-scored against gold answer, while FinePDFs~\cite{kydlicek2025finepdfs} and RoMemes-OCR use Character Error Rate (CER), Word Error Rate (WER), and Average Normalized Levenshtein Similarity (ANLS). \textit{Grounding} with Pixmo-Count and Pixmo-Points~\cite{deitke2025molmo} measures counting and pointing accuracy, with an LLM used only to parse the model's free-form answer before comparing against the gold value. 

As can be seen in Table~\ref{eval-benchmarks}, about half of the benchmarks have native Romanian text; the rest were translated as described in Section~\ref{sec:translation}. The full scoring protocol — per-benchmark metrics, judge models, and prompts — is given in Appendix~\ref{sec:appendix-eval-protocol}.

\begin{table} [hbt]
  \centering
  \begin{tabular}{llr}
    \hline
    \textbf{Task} & \textbf{Dataset} & \textbf{\# Samples} \\
    \hline
    \multirow{3}{*}{\shortstack[l]{General\\Understanding}}
        & MMBench           & 4{,}876  \\
        & MMStar            & 1{,}500  \\
        & SeedBench2        & 23{,}279 \\
    \hline
    \multirow{2}{*}{\shortstack[l]{Knowledge \&\\Reasoning}}
        & MMMU              & 900     \\
        & MME               & 2{,}374 \\
    \hline
    \multirow{4}{*}{Cultural}
        & CVQA$^{\dagger}$       & 302     \\
        & ALM-Bench$^{\dagger}$  & 226     \\
        & RoMemes$^{\dagger}$    & 1{,}848 \\
        & HoraVQA$^{\dagger}$  & 580     \\
    \hline
    \multirow{5}{*}{\shortstack[l]{Generation \&\\Open-ended}}
        & Flickr30k-Caption          & 6{,}357 \\
        & Flickr30k-QA               & 6{,}357 \\
        & LLaVA-Wild                 & 60      \\
        & AyaVisionBench$^{\dagger}$ & 135     \\
        & m-WildVision$^{\dagger}$   & 500     \\
    \hline
    \multirow{3}{*}{\shortstack[l]{OCR \&\\Documents}}
        & CoSyn                  & 7{,}966 \\
        & FinePDFs$^{\dagger}$   & 1{,}254 \\
        & RoMemes-OCR$^{\dagger}$ & 462    \\
    \hline
    \multirow{2}{*}{Grounding}
        & Pixmo-Count       & 527 \\
        & Pixmo-Points      & 337 \\
    \hline
    \textbf{Total} & & \textbf{59{,}840} \\
    \hline
  \end{tabular}
  \caption{\label{eval-benchmarks}
    Evaluation benchmarks grouped by the capability they probe. Datasets marked with $\dagger$ have Romanian question/answer text in their released form and are used as-is; unmarked datasets were translated from English by us. For two of the marked datasets (ALM-Bench and m-WildVision), the text is native Romanian but the in-image text was additionally translated by us. The translation and diacritic-restoration pipeline is described in Section~\ref{sec:translation}.}
\end{table}

\section{Experiments and Analysis}
In this section, we investigate (i) the impact of the collected corpora, (ii) which components — language or vision backbone — matter most, (iii) how important language-specific OCR data is, and (iv) the impact of translated images.

\subsection{Main Results}

\begin{figure}[h]
  \centering
  \includegraphics[width=0.8\columnwidth]{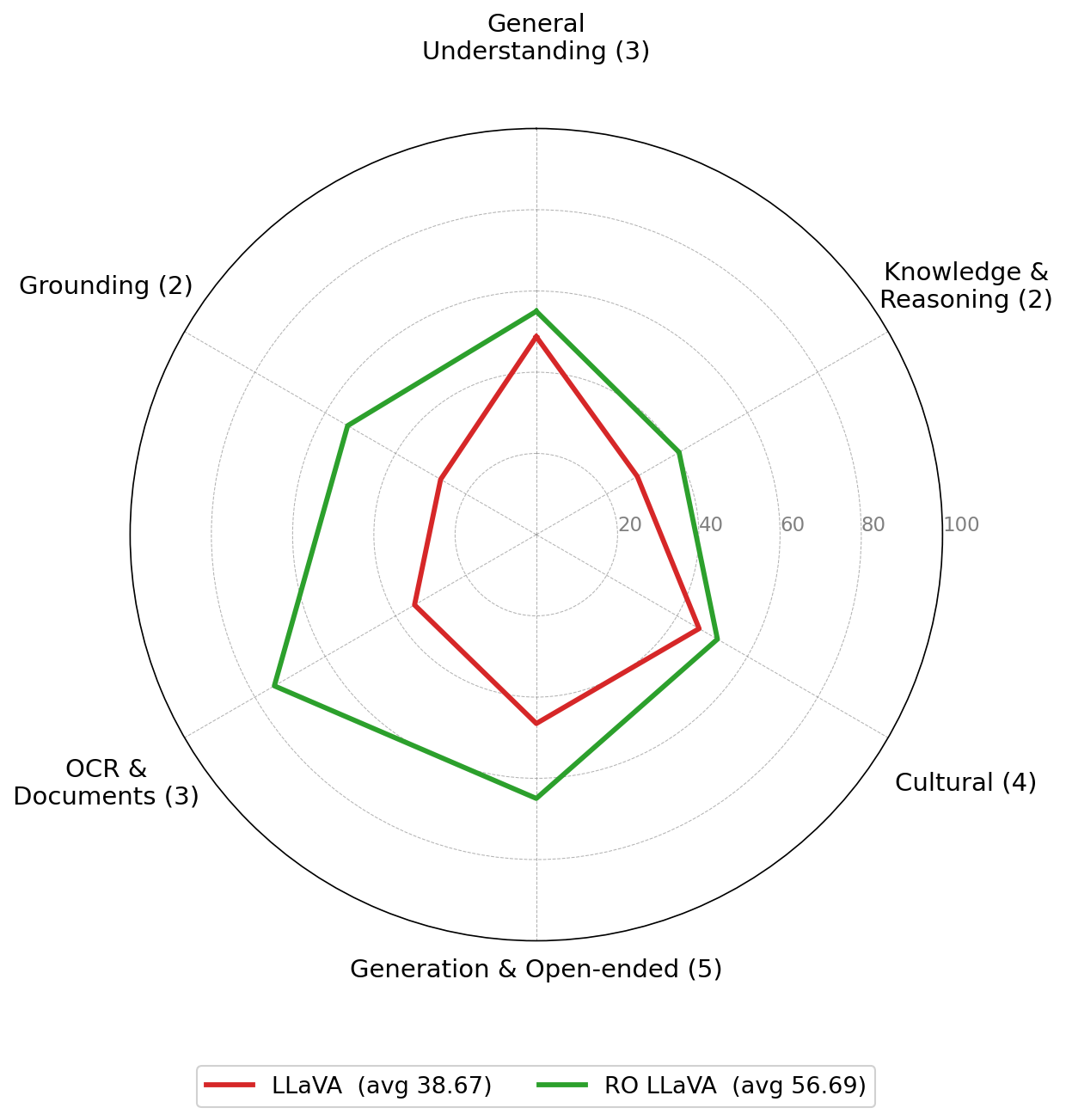}
  \caption{Performance comparison between the original LLaVA model and its Romanian adaptation. We note the stronger performance of RoLLaVA across each category.}
  \label{fig:llava_base_vs_sft}
\end{figure}

First, we evaluate the extent to which our collected data actually improves performance in Romanian. To this end, we perform supervised fine-tuning on LLaVA-NeXT-Llama3-8B\footnote{llava-hf/llama3-llava-next-8b-hf} with our collected data, leading to \textit{RoLLaVA} models. Figure~\ref{fig:llava_base_vs_sft} shows the clear improvement over the base model, with RoLLaVA improving across all categories, building an almost 20-point gap.

Furthermore, we show that this improvement is not merely a byproduct of the architecture choice, but generalizes across vision encoders, language backbones, and model  architectures. The results summarized in Table~\ref{tab:summary} stand as testament to that, with full results presented in Appendix~\ref{sec:appendix-additional-results}.

\begin{table}[hb]
  \centering
  \begin{tabular}{l|rr}
    \hline
    \textbf{Model} & \textbf{OG} & \textbf{RO}\\
    \hline
    LLaVA-NeXT-Llama3-8B & 38.67 & 56.05 \\
    \hline
    Qwen2-VL-2B & 40.56 & 57.88 \\
    Qwen2-VL-7B & 57.49 & --- \\
    \hline
    Qwen2.5-VL-3B & 52.99 & 61.68 \\
    Qwen2.5-VL-7B & 60.59 & --- \\
    \hline
    Qwen3-VL-2B & 51.31 & 62.65 \\
    Qwen3-VL-4B & 61.35 & --- \\
    Qwen3-VL-8B & 62.69 & --- \\
    \hline
    Gemma3-4B & 52.36 & 57.14 \\
    Gemma3-12B & 59.49 & --- \\
    \hline
  \end{tabular}
  \caption{\label{tab:summary}
    Average performance, original model (\textbf{OG}, all instruct versions) versus Romanian-adapted variant (\textbf{RO}).}
\end{table}

\subsection{HoraVQA Evaluation}

Results on the curated HoraVQA benchmark are summarized in Table~\ref{tab:horavqa-overall}. Across multiple architectures, RoVLMs consistently outperform their original counterparts, with particularly substantial gains observed for LLaVA-NeXT, Qwen2-VL, and Qwen3-VL. Improvements for Qwen2.5-VL and Gemma3 are comparatively smaller, though still consistently positive. Comprehensive results are provided in Appendix~\ref{sec:appendix-additional-results}.

\begin{table}
  \centering
  \footnotesize
  \begin{tabular}{lc}
    \hline
    \textbf{Model} & \textbf{HoraVQA Score} \\
    \hline
    LLaVA-NeXT-Llama3-8B & 40.60 \\
    \textit{RO-LLaVA-NeXT-Llama3-8B} & \textit{47.78} \\
    \hline
    Qwen2-VL-2B-Instruct & 46.24 \\
    \textit{RO-Qwen2-VL-2B} & \textit{56.09} \\
    Qwen2-VL-7B-Instruct & 58.09 \\
    \hline
    Qwen2.5-VL-3B-Instruct & 54.19 \\
    \textit{RO-Qwen2.5-VL-3B} & \textit{56.36} \\
    Qwen2.5-VL-7B-Instruct & 61.00 \\
    \hline
    Qwen3-VL-2B-Instruct & 50.31 \\
    \textit{RO-Qwen3-VL-2B} & \textit{54.00} \\
    Qwen3-VL-4B-Instruct & 58.98 \\
    Qwen3-VL-8B-Instruct & 60.81 \\
    \hline
    Gemma3-4B-it & 52.19 \\
    \textit{RO-Gemma3-4B} & \textit{52.84} \\
    Gemma3-12B-it & 58.91 \\
    \hline
  \end{tabular}
  \caption{\label{tab:horavqa-overall}  Overall HoraVQA performance per model.}
\end{table}

\subsection{Influence of Language Backbone}

To investigate the importance of a specialized language backbone, we start from the LLaVA-NeXT architecture and set the language backbones to Llama3-8B-Instruct\footnote{meta-llama/Meta-Llama-3-8B} and its Romanian counterpart\footnote{OpenLLM-Ro/RoLlama3-8b-Instruct-DPO}. We kept the same vision encoder and re-initialized the vision adapter (randomly, same exact setup for both models). Afterward, we use the same training setup as before. 

In this case, the influence is negligible (56.05 RoLlama3 vs 55.92 for Llama3, see Appendix~\ref{sec:appendix-additional-results}) despite RoLlama3 showing significant improvements on linguistic tasks~\cite{masala2024vorbești}. This outcome can be attributed to several factors: (i) Llama3 had already a good understanding of Romanian, (ii) our ~3.1M training samples provide provided substantial capacity for fine-tuning for Romanian understanding and generation, (iii) RoLlaMA3 was trained using a maximum sequence length of only 1024, which is exceeded in most of cases for multimodal tasks (i.e., an image of 512x512 already uses 1152 image tokens for LLaVA-NeXT).

Consistent with previous results, we observe that training in this setup (starting from a pre-trained vision encoder with a randomly initialized vision adapter and a text-only LLM) largely matches training from an existing checkpoint (where the adapter and LLM are already fine-tuned on English data): 56.69 versus 56.05.

\subsection{Influence of Vision Backbone}

\begin{figure}[h]
  \centering
  \includegraphics[width=0.85\columnwidth]{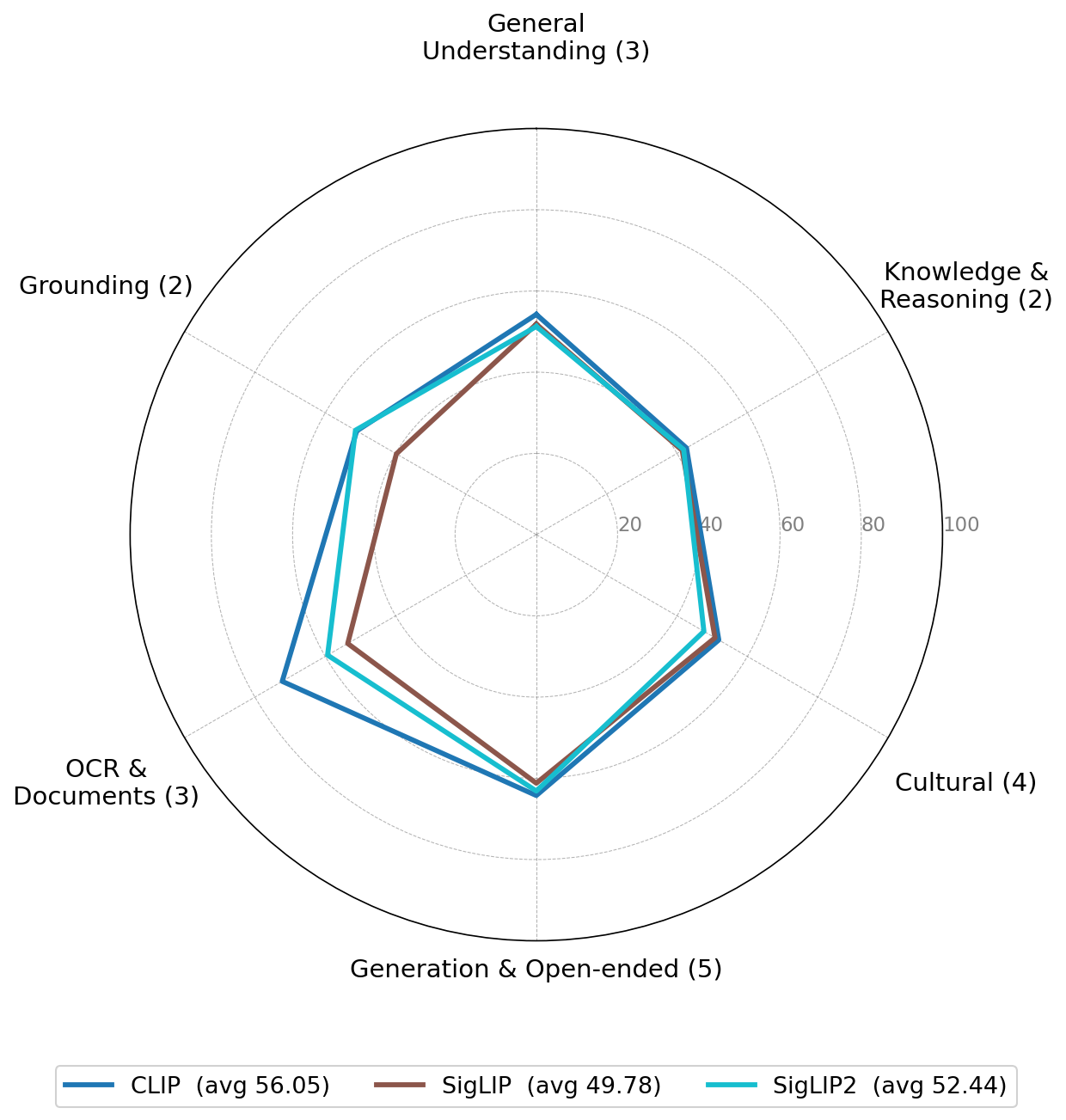}
  \caption{Performance comparison between CLIP, SigSLIP, and SigLIP2 visual backbones. Note the stronger overall performance of the CLIP backbone, especially on \textit{OCR \& Documents} category.}
  \label{fig:clip_vs_siglip_vs_siglip2}
\end{figure}

For the vision encoder, we experiment with three different variants: CLIP\footnote{openai/clip-vit-large-patch14-336} (as in LLaVA-NeXT), SigLIP\footnote{google/siglip-so400m-patch14-384}, and SigLIP2\footnote{google/siglip2-so400m-patch14-384}. Figure~\ref{fig:clip_vs_siglip_vs_siglip2} highlights several notable findings. SigLIP2, with its stronger emphasis on large-scale multilingual alignment, consistently outperforms SigLIP across all categories except the Cultural group, where it trails by a small margin. In the remaining categories, SigLIP2 either matches or surpasses SigLIP. However, neither SigLIP nor SigLIP2 shows improvements over CLIP; in particular, both models are severely outperformed by CLIP on the \textit{OCR\&Documents} category.

A plausible explanation for CLIP outperforming SigLIP and SigLIP2 as a vision encoder for Romanian OCR tasks is that CLIP preserves more locally discriminative, high-frequency visual structure, which is critical for exact character transcription. In contrast, SigLIP and especially SigLIP2 optimize more strongly for semantic robustness, multilingual alignment, localization, and dense feature consistency, which may improve retrieval and grounding but can suppress the fragile local visual details needed for autoregressive OCR decoding~\cite{zhai2023sigmoid,tschannen2025siglip}. This creates an objective mismatch: OCR requires literal visual fidelity, while modern contrastive VLM encoders increasingly optimize toward semantic abstraction and invariance. Similarly, \citet{zhang2026penguin} argue that contrastive encoders such as CLIP and SigLIP tend to enforce coarse category-level invariances that may suppress fine-grained visual cues required for downstream dense visual reasoning tasks. 

\subsection{Influence of OCR Data}

\begin{figure}[h]
  \centering
  \includegraphics[width=0.85\columnwidth]{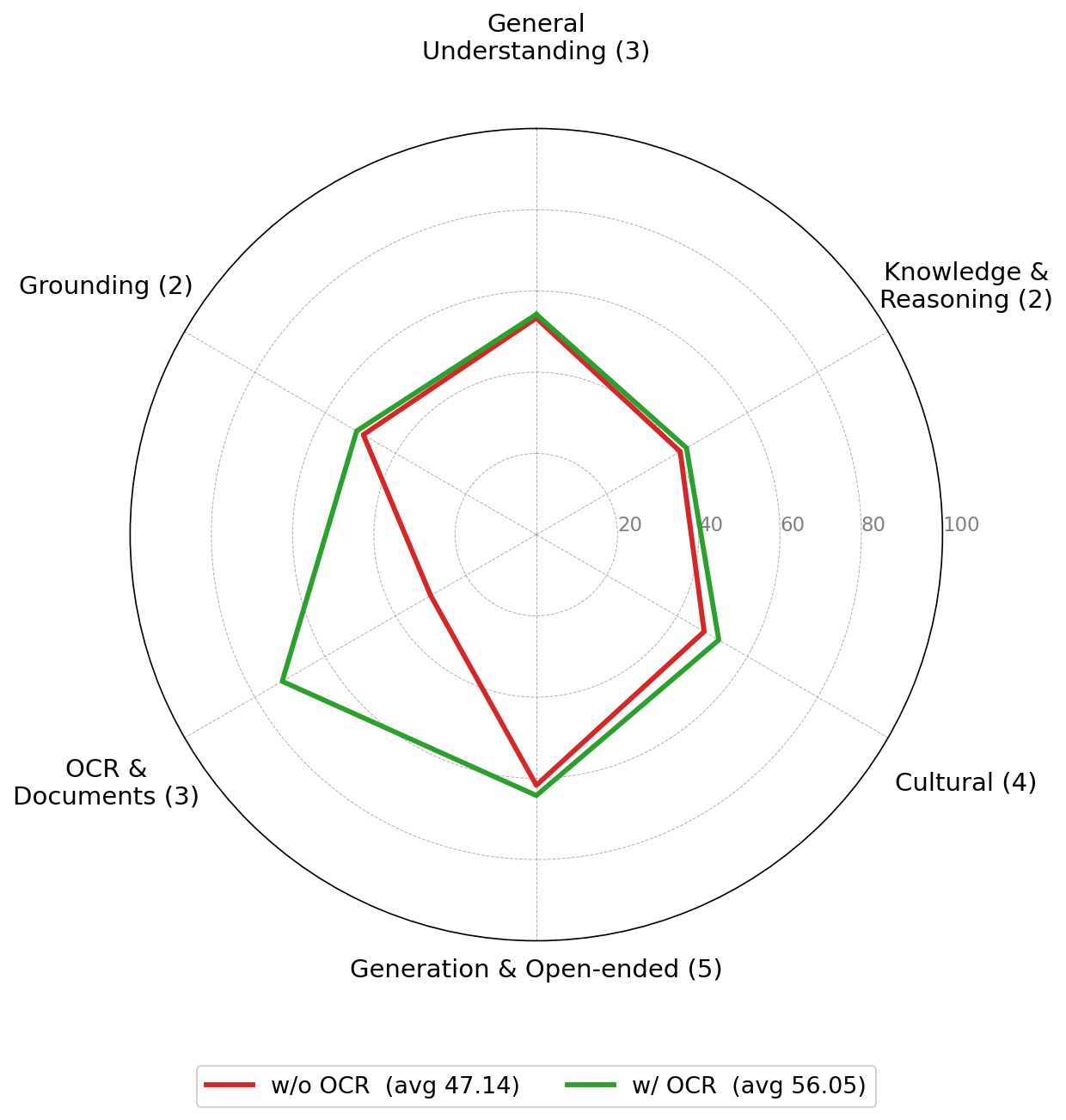}
  \caption{Performance comparison between the model trained on full data (green line) and without OCR data (red line). Note that performance decreases in all categories.}
  \label{fig:ocr_ablation.png}
\end{figure}

In this section, we evaluate the influence of OCR data on downstream performance. Removing OCR data leads to lower performance across the board, not only on OCR-heavy evaluation (see red lines in Figure~\ref{fig:ocr_ablation.png}). Even in other categories (e.g., Cultural), certain evaluation samples may require text recognition within images. In such cases, the model needs to rely on its native OCR capabilities.

\subsection{Influence of Translated Images}

\begin{figure}[h]
  \centering
  \includegraphics[width=0.85\columnwidth]{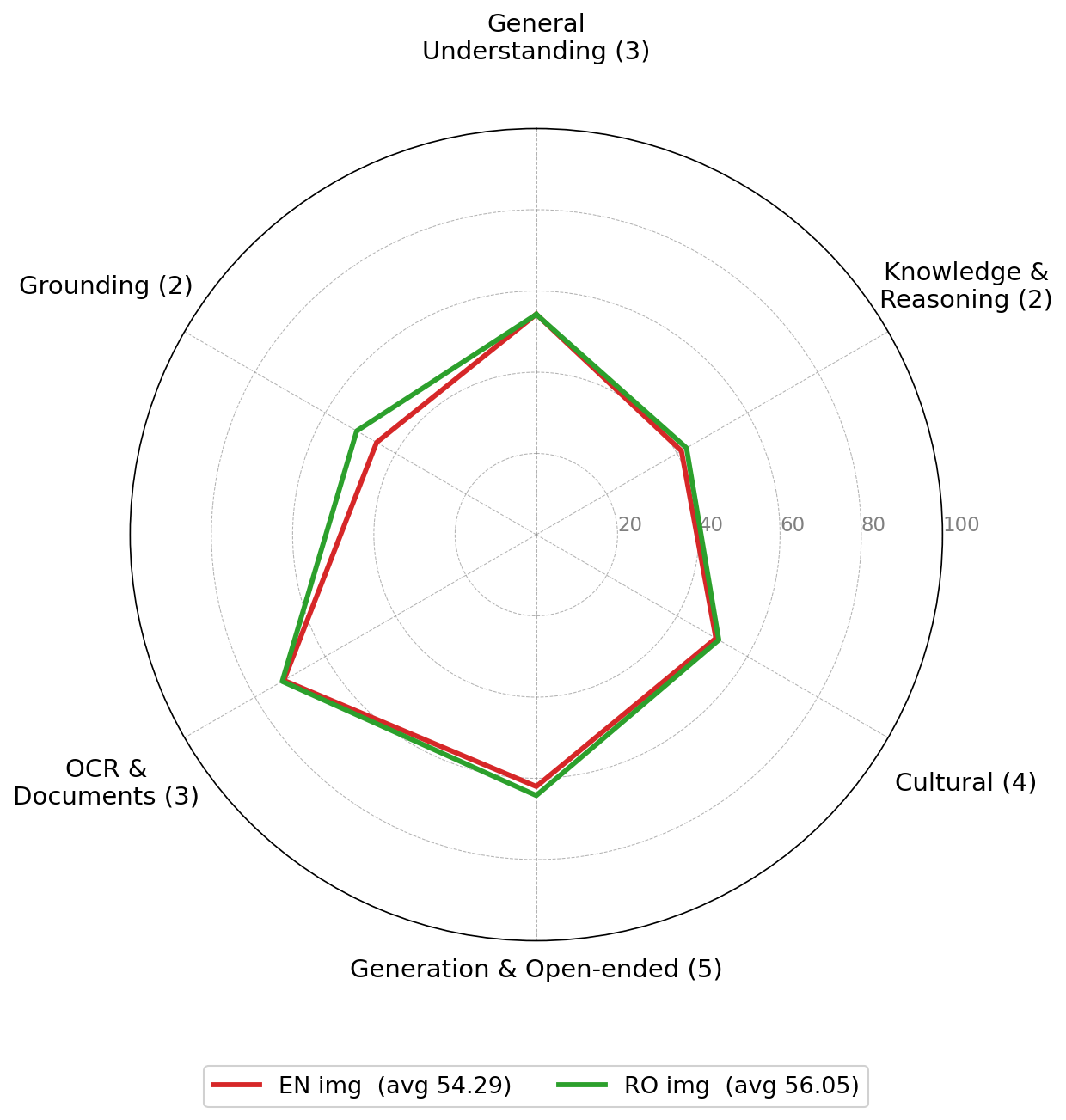}
  \caption{Performance comparison between the model trained on full Romanian images (green line) and without English images (except raw OCR data). Note the \textit{Grounding} and \textit{average} performance drop when using English images.}
  \label{fig:en_vs_ro_img.png}
\end{figure}

Finally, we are interested in the importance of images having text in the targeted language vs English sources. Results in Figure~\ref{fig:en_vs_ro_img.png} reveal that, using translated images in the target language slightly increases the performance, with an average of 56.05 vs 54.29 for the baseline that employs native English images. Out of all categories, the largest performance gap can be observed for \textit{Grounding} - around 5.6 points, followed by \textit{Generation \& Open-ended} categories with around 2.2 points.

\section{Conclusion}

The dominant trajectory in multimodal modeling has been to scale a single generalist VLM across as many languages and tasks as possible. Our results argue that this is not the whole picture: specialized, language-targeted adaptation still pays for itself, even against models one size class larger. By combining translated multimodal supervision, in-image text translation, code-level regeneration of synthetic charts, and a substantial OCR/document mix, RoVLMs match or surpass English-only counterparts and several larger generalist baselines across 19 Romanian benchmarks. Together with HoraVQA, our native-speaker cultural benchmark, this gives the community both a recipe and an evaluation suite for Romanian multimodal work.

Beyond the raw numbers, our ablations show \textit{why} and \textit{how} specialized adaptation is worth doing. A Romanian-tuned text backbone alone changes little; what moves the needle is data composition --- particularly OCR supervision, which improves every task category, not just text-rich ones --- and the choice of vision encoder, where CLIP outperforms the multilingual SigLIP2 on OCR-sensitive splits despite SigLIP2's broader language coverage. The implication is that ``multilingual'' at the representation level does not automatically deliver grounded, OCR-faithful, culturally aware multimodal behavior. Specialized data and specialized benchmarks remain necessary.

\section*{Limitations}

Although human-validated, most of the training and evaluation data originates from translated sources, largely due to the scarcity of high-quality, language-specific instruction datasets. Errors may arise both from translation of the textual instruction components (e.g., question–answer pairs) and from in-image text translation. Our in-image translation pipeline still leaves residual untranslated text in complex layouts, while the CoSyn regeneration strategy is applicable only to synthetic data with available source material. 

Our findings are established on Romanian alone; whether the same adaptation recipe transfers to other low-resource European languages is an empirical question we do not tackle in this work.

A key limitation of our approach is that the models are trained primarily on existing and translated datasets, over which we have limited control. As a result, biases present in the original training data and inherited from upstream models or even from the translation models used, may persist and propagate into our system outputs. These biases can affect representation, language use, and downstream predictions. In addition, the current models do not incorporate dedicated safety mechanisms, meaning they may generate harmful, misleading, or otherwise unsafe content.




\bibliography{custom}

\appendix

\section{Additional Results}
\label{sec:appendix-additional-results}

\begin{figure}[h]
  \centering
  \includegraphics[width=0.85\columnwidth]{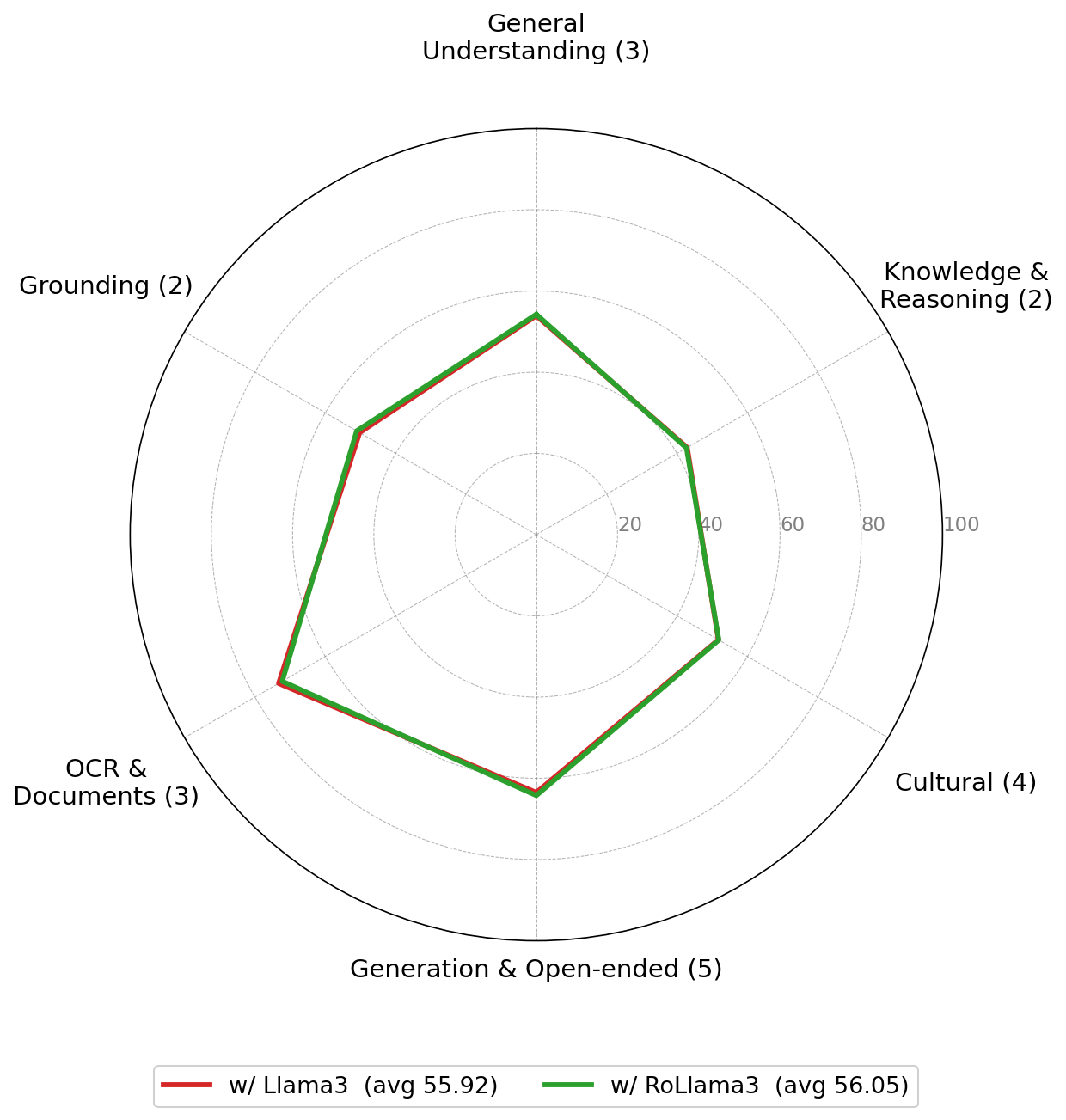}
  \caption{Performance comparison between Llama3 and RoLlama3 language backbone. Note that both backbones perform similarly.}
  \label{fig:llama_vs_rollama}
\end{figure}

Language backbone influence results are presented in Figure~\ref{fig:llama_vs_rollama}. 

\begin{figure*}[h]
  \includegraphics[width=\textwidth]{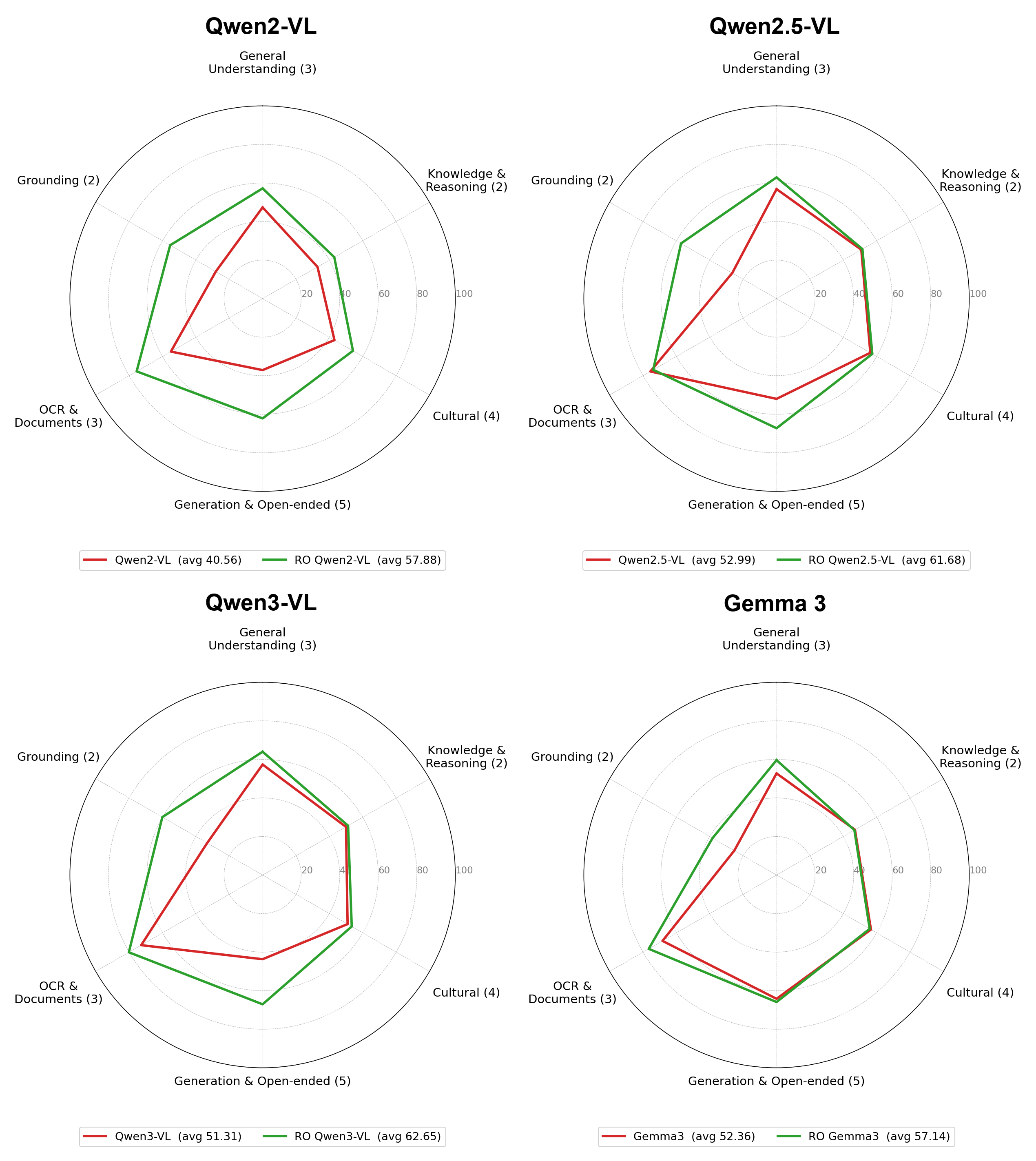}
  \caption{Performance comparison between the original models and the Romanian adaptations across multiple architectures. Note the stronger performance of RO variant across each architecture. For newer and stronger base models, the gains less but still clearly visible.}
  \label{fig:multiple_architectures.png}
\end{figure*}

The results for each category across multiple architectures are presented in the Figure~\ref{fig:multiple_architectures.png}. There is a clear improvement across all considered architectures, with the performance gap closing with stronger models (e.g. Qwen3-VL vs Qwen2-VL).

Full results of the evaluated models, with per benchmark scores are presented in Table~\ref{tab:full-results}.

\begin{sidewaystable*}
  \centering
  \footnotesize
  \setlength{\tabcolsep}{4pt}
  \resizebox{\textheight}{!}{%
  \begin{tabular}{l|cc|ccc|cc|cccc|ccccc|ccc|cc}
    \hline
     & \multicolumn{2}{c}{\textit{Aggregate}} & \multicolumn{3}{c}{\textit{General Understanding}} & \multicolumn{2}{c}{\textit{Knowledge \& Reasoning}} & \multicolumn{4}{c}{\textit{Cultural}} & \multicolumn{5}{c}{\textit{Generation \& Open-ended}} & \multicolumn{3}{c}{\textit{OCR \& Documents}} & \multicolumn{2}{c}{\textit{Grounding}} \\
    \textbf{Model} & \rotatebox{90}{Micro} & \rotatebox{90}{Macro} & \rotatebox{90}{MMBench} & \rotatebox{90}{MMStar} & \rotatebox{90}{SeedBench2} & \rotatebox{90}{MMMU} & \rotatebox{90}{MME} & \rotatebox{90}{CVQA} & \rotatebox{90}{ALM-Bench} & \rotatebox{90}{RoMemes} & \rotatebox{90}{RoCultVLM} & \rotatebox{90}{RoFlickr30k-Caption} & \rotatebox{90}{RoFlickr30k-QA} & \rotatebox{90}{LLaVA-Wild} & \rotatebox{90}{AyaVisionBench} & \rotatebox{90}{m-WildVision} & \rotatebox{90}{RoCosyn} & \rotatebox{90}{RoFinepdfs} & \rotatebox{90}{RoMemes OCR} & \rotatebox{90}{PixmoCount} & \rotatebox{90}{PixmoPoints} \\
    \hline
    llava-v1.6-mistral-7b-4bit-RoVQA-lora & 34.84 & 33.61 & 49.69 & 33.54 & 52.96 & 26.67 & 46.24 & 21.52 & 44.87 & 11.16 & 25.50 & 74.33 & 70.34 & 24.01 & 30.67 & 32.46 & 31.51 & 0.77 & 35.62 & 39.66 & 10.39 \\
    Llama-3.2-11B-Vision-Instruct-RoVQA & 50.18 & 47.17 & 58.36 & 45.17 & 60.66 & 39.44 & 45.50 & 70.86 & 69.56 & 36.20 & 55.16 & 80.45 & 83.10 & 42.75 & 43.85 & 45.22 & 46.60 & 7.27 & 64.87 & 48.01 & 10.39 \\
    LLaVA-NeXT-Mistral-7B & 38.18 & 36.54 & 48.37 & 30.52 & 51.85 & 31.67 & 39.15 & 55.63 & 52.88 & 30.25 & 37.90 & 66.68 & 55.71 & 29.58 & 26.15 & 39.18 & 32.29 & 1.99 & 39.66 & 45.54 & 10.39 \\
    LLaVA-NeXT-Llama3-8B & 41.02 & 38.67 & 60.91 & 37.34 & 47.93 & 33.22 & 24.16 & 57.62 & 47.43 & 39.45 & 40.60 & 67.60 & 61.25 & 29.58 & 32.52 & 41.56 & 36.33 & 3.52 & 64.01 & 44.02 & 10.39 \\
    \textit{RO-LLaVA-NeXT-Llama3-8B} & \textit{58.29} & \textit{56.69} & 67.72 & 43.32 & 53.92 & 34.67 & 46.49 & 59.60 & 55.88 & 42.66 & 47.78 & 84.71 & 85.53 & 50.59 & 46.59 & 57.34 & 57.83 & 78.13 & 87.46 & 57.12 & 50.16 \\
    \hline
    Qwen2-VL-2B-Instruct & 41.39 & 40.56 & 51.47 & 37.94 & 52.54 & 34.56 & 31.22 & 57.62 & 37.57 & 30.77 & 46.24 & 64.55 & 45.64 & 21.17 & 24.44 & 29.66 & 41.01 & 37.53 & 86.34 & 45.73 & 10.39 \\
    \textit{RO-Qwen2-VL-2B} & \textit{59.05} & \textit{57.88} & 65.25 & 40.50 & 65.84 & 36.78 & 49.10 & 65.56 & 60.88 & 34.00 & 56.09 & 83.69 & 82.92 & 42.49 & 45.70 & 55.92 & 57.22 & 85.50 & 83.74 & 63.76 & 47.03 \\
    Qwen2-VL-7B-Instruct & 59.13 & 57.49 & 68.42 & 52.47 & 66.30 & 45.78 & 65.48 & 72.19 & 64.64 & 55.16 & 58.09 & 70.88 & 71.88 & 34.75 & 51.41 & 56.06 & 55.40 & 79.57 & 90.48 & 54.08 & 10.39 \\
    \hline
    Qwen2.5-VL-3B-Instruct & 54.56 & 52.99 & 62.00 & 47.50 & 60.95 & 40.56 & 61.00 & 64.90 & 55.53 & 50.15 & 54.19 & 71.04 & 60.43 & 37.86 & 43.11 & 47.64 & 56.04 & 79.45 & 91.29 & 42.69 & 10.39 \\
    \textit{RO-Qwen2.5-VL-3B} & \textit{62.81} & \textit{61.68} & 69.97 & 50.42 & 68.30 & 41.22 & 61.68 & 70.53 & 67.12 & 35.51 & 56.36 & 83.46 & 83.44 & 53.88 & 55.56 & 59.90 & 67.75 & 85.96 & 67.93 & 69.64 & 44.82 \\
    Qwen2.5-VL-7B-Instruct & 62.84 & 60.59 & 72.91 & 56.25 & 69.44 & 45.78 & 61.18 & 72.85 & 67.96 & 54.19 & 61.00 & 70.92 & 74.87 & 51.18 & 59.78 & 65.10 & 66.45 & 81.48 & 91.07 & 61.11 & 10.53 \\
    \hline
    Qwen3-VL-2B-Instruct & 51.51 & 51.31 & 62.69 & 45.92 & 63.38 & 38.33 & 61.59 & 57.95 & 48.72 & 46.68 & 50.31 & 70.09 & 30.59 & 29.89 & 43.04 & 44.76 & 48.63 & 78.62 & 91.04 & 56.36 & 10.09 \\
    \textit{RO-Qwen3-VL-2B} & \textit{63.36} & \textit{62.65} & 71.90 & 50.73 & 69.29 & 40.22 & 62.19 & 61.92 & 60.97 & 36.71 & 54.00 & 83.80 & 85.70 & 50.40 & 55.33 & 60.08 & 64.07 & 86.85 & 89.54 & 65.28 & 54.89 \\
    Qwen3-VL-4B-Instruct & 63.07 & 61.35 & 74.07 & 55.28 & 71.18 & 49.33 & 75.12 & 67.55 & 57.96 & 55.84 & 58.98 & 70.15 & 90.48 & 38.29 & 69.04 & 61.18 & 62.63 & 80.36 & 89.92 & 60.15 & 10.81 \\
    Qwen3-VL-8B-Instruct & 64.84 & 62.69 & 76.39 & 56.92 & 72.53 & 50.44 & 77.05 & 71.52 & 61.73 & 54.94 & 60.81 & 70.41 & 95.49 & 45.10 & 72.44 & 64.20 & 64.07 & 80.31 & 89.47 & 57.12 & 11.04 \\
    \hline
    Gemma3-4B-it & 55.53 & 52.36 & 59.13 & 41.49 & 57.55 & 36.67 & 57.32 & 64.90 & 65.97 & 43.24 & 52.19 & 70.93 & 81.66 & 54.84 & 52.81 & 60.80 & 48.40 & 67.12 & 89.47 & 40.42 & 10.21 \\
    \textit{RO-Gemma3-4B} & \textit{59.54} & \textit{57.14} & 69.96 & 46.01 & 62.83 & 38.67 & 54.62 & 64.24 & 65.40 & 40.78 & 52.84 & 84.35 & 84.74 & 55.71 & 47.04 & 57.60 & 59.06 & 84.35 & 86.33 & 51.80 & 24.87 \\
    Gemma3-12B-it & 63.23 & 59.49 & 70.43 & 51.86 & 65.59 & 46.67 & 70.22 & 76.16 & 79.30 & 52.33 & 58.91 & 71.55 & 87.15 & 74.53 & 67.63 & 70.66 & 55.40 & 70.78 & 79.52 & 42.31 & 10.39 \\
    \hline
    InternVL3\_5-2B & 49.79 & 49.00 & 62.46 & 51.86 & 61.85 & 42.11 & 45.11 & 49.34 & 47.48 & 47.02 & 42.55 & 68.78 & 53.66 & 20.30 & 42.30 & 38.82 & 49.12 & 76.16 & 89.92 & 46.87 & 10.29 \\
    \hline
  \end{tabular}
  }
  \caption{\label{tab:full-results}
    Per-benchmark results across all evaluated VLMs. Columns are grouped by capability (general understanding, knowledge \& reasoning, cultural, generation \& open-ended, OCR \& documents, grounding). Micro is the mean over individual benchmarks; Macro is the mean over capability groups.}
\end{sidewaystable*}

Full results on HoraVQA are presented in Table~\ref{tab:horavqa-categories}. We note that regarding format, open-ended question are significantly harder to multi-choice questions, with an almost 25 point gap. Similarly, models are weaker on culturally-grounded items with an 11 point difference. Topic-level scores span roughly fourteen points and fall into three tiers. The strongest results — People (56.52), Traditions (55.28), and Daily life (55.06) — involve visually generic content that benefits from web-scale pretraining regardless of cultural origin. A middle tier (Recent History 52.93, Transport 51.45, Food\&Drinks 49.61, Landmarks 49.50) covers concepts that are visually identifiable but require linking a recognizable cue to a specific Romanian referent, and is therefore bottlenecked by named-entity knowledge of Romania. The weakest tier — Sports (48.14) and especially Arts (43.00) — is dominated by knowledge-heavy categories that demand fine-grained discrimination of specific athletes, events, artworks, and styles, areas that are both more demanding visually and chronically underrepresented for Romanian culture in mainstream VLM training data.

\begin{sidewaystable*}
  \centering
  \footnotesize
  \setlength{\tabcolsep}{4pt}
  \resizebox{\textheight}{!}{%
  \begin{tabular}{l|c|cc|cc|ccccccccc}
    \hline
     & \multicolumn{1}{c}{\textit{Overall}} & \multicolumn{2}{c}{\textit{Format}} & \multicolumn{2}{c}{\textit{Cultural}} & \multicolumn{9}{c}{\textit{Topic}} \\
    \textbf{Model} & \rotatebox{90}{All} & \rotatebox{90}{MCQ} & \rotatebox{90}{Open} & \rotatebox{90}{Cult} & \rotatebox{90}{Non-Cult} & \rotatebox{90}{Arts} & \rotatebox{90}{Recent History} & \rotatebox{90}{Daily} & \rotatebox{90}{Traditions} & \rotatebox{90}{Food\&Drinks} & \rotatebox{90}{Landmarks} & \rotatebox{90}{People} & \rotatebox{90}{Sports} & \rotatebox{90}{Transport} \\
    \hline
    llava-v1.6-mistral-7b-4bit-RoVQA-lora & 25.50 & 26.65 & 23.06 & 27.63 & 22.71 & 21.28 & 19.20 & 24.46 & 31.11 & 22.24 & 27.16 & 23.87 & 30.30 & 23.26 \\
    Llama-3.2-11B-Vision-Instruct-RoVQA & 55.16 & 62.69 & 39.19 & 54.07 & 56.57 & 41.03 & 61.60 & 63.93 & 59.63 & 55.14 & 54.94 & 56.13 & 56.97 & 43.91 \\
    LLaVA-NeXT-Mistral-7B & 37.90 & 46.70 & 19.25 & 34.04 & 42.95 & 26.92 & 36.80 & 43.39 & 45.31 & 39.63 & 38.02 & 34.19 & 24.55 & 35.65 \\
    LLaVA-NeXT-Llama3-8B & 40.60 & 50.25 & 20.16 & 37.48 & 44.70 & 22.56 & 49.60 & 46.61 & 53.21 & 40.37 & 39.57 & 30.00 & 35.76 & 36.30 \\
    \textit{RO-LLaVA-NeXT-Llama3-8B} & \textit{47.78} & 56.35 & 29.62 & 40.64 & 57.13 & 38.46 & 54.00 & 48.75 & 54.81 & 47.57 & 43.52 & 44.84 & 54.24 & 51.52 \\
    \hline
    Qwen2-VL-2B-Instruct & 46.24 & 58.38 & 20.54 & 43.71 & 49.56 & 40.77 & 46.80 & 52.50 & 49.63 & 39.81 & 47.90 & 63.23 & 41.52 & 38.04 \\
    \textit{RO-Qwen2-VL-2B} & \textit{56.09} & 65.99 & 35.11 & 48.12 & 66.53 & 53.59 & 53.60 & 68.21 & 63.09 & 59.16 & 49.88 & 60.32 & 40.61 & 55.43 \\
    Qwen2-VL-7B-Instruct & 58.09 & 66.75 & 39.73 & 52.43 & 65.50 & 51.28 & 71.20 & 61.25 & 62.59 & 47.20 & 58.09 & 71.94 & 54.24 & 63.70 \\
    \hline
    Qwen2.5-VL-3B-Instruct & 54.19 & 60.91 & 39.95 & 52.67 & 56.18 & 50.77 & 54.40 & 61.96 & 56.42 & 42.99 & 56.54 & 66.45 & 56.97 & 51.09 \\
    \textit{RO-Qwen2.5-VL-3B} & \textit{56.36} & 64.47 & 39.19 & 51.09 & 63.27 & 51.28 & 49.60 & 53.93 & 58.64 & 57.66 & 56.91 & 71.29 & 45.76 & 55.87 \\
    Qwen2.5-VL-7B-Instruct & 61.00 & 66.50 & 49.35 & 54.86 & 69.04 & 53.85 & 56.00 & 68.57 & 63.70 & 55.51 & 59.38 & 73.55 & 61.52 & 65.43 \\
    \hline
    Qwen3-VL-2B-Instruct & 50.31 & 58.12 & 33.76 & 43.80 & 58.84 & 42.05 & 51.20 & 55.54 & 49.63 & 47.66 & 49.69 & 57.74 & 49.70 & 55.43 \\
    \textit{RO-Qwen3-VL-2B} & \textit{54.00} & 61.17 & 38.82 & 46.69 & 63.59 & 51.79 & 59.60 & 53.75 & 54.69 & 57.94 & 50.19 & 62.26 & 45.15 & 56.96 \\
    Qwen3-VL-4B-Instruct & 58.98 & 68.78 & 38.23 & 52.31 & 67.73 & 50.51 & 62.00 & 66.96 & 57.90 & 54.30 & 58.77 & 69.03 & 52.73 & 66.09 \\
    Qwen3-VL-8B-Instruct & 60.81 & 70.56 & 40.16 & 53.13 & 70.88 & 48.46 & 69.20 & 63.39 & 60.25 & 64.77 & 56.30 & 70.32 & 70.30 & 58.04 \\
    \hline
    Gemma3-4B-it & 52.19 & 58.88 & 38.01 & 49.36 & 55.90 & 41.54 & 50.00 & 51.25 & 54.57 & 57.48 & 51.30 & 52.90 & 52.12 & 49.78 \\
    \textit{RO-Gemma3-4B} & \textit{52.84} & 60.66 & 36.29 & 49.18 & 57.65 & 40.77 & 47.20 & 51.25 & 55.80 & 58.04 & 50.74 & 56.45 & 53.64 & 55.22 \\
    Gemma3-12B-it & 58.91 & 63.96 & 48.23 & 57.69 & 60.52 & 45.13 & 68.40 & 58.21 & 68.64 & 62.99 & 53.70 & 62.58 & 51.52 & 60.87 \\
    \hline
    InternVL3\_5-2B & 42.55 & 52.03 & 22.47 & 33.40 & 54.54 & 44.87 & 45.20 & 52.14 & 50.62 & 32.06 & 37.96 & 46.77 & 36.97 & 55.00 \\
    \hline
    \textbf{Average} & \textbf{51.03} & \textbf{58.94} & \textbf{34.27} & \textbf{46.44} & \textbf{57.04} & \textbf{43.00} & \textbf{52.93} & \textbf{55.06} & \textbf{55.28} & \textbf{49.61} & \textbf{49.50} & \textbf{56.52} & \textbf{48.14} & \textbf{51.45} \\
    \hline
  \end{tabular}
  }
  \caption{\label{tab:horavqa-categories}
    HoraVQA per-category scores across all evaluated VLMs. Columns are grouped by format (MCQ vs. open-ended), cultural-vs-non-cultural split, and topic category.}
\end{sidewaystable*}






\section{Translation Process}
\label{sec:appendix-translation-process}

\begin{figure*}[h]
  \includegraphics[width=\textwidth]{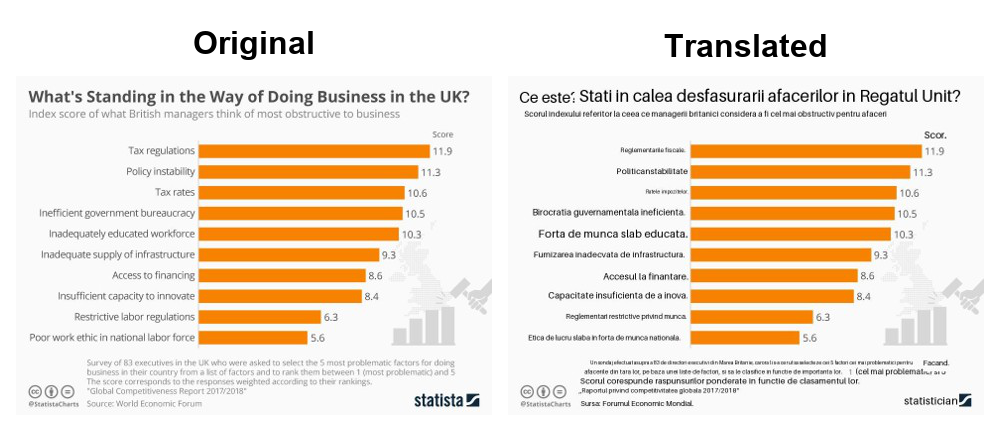}
  \caption{Example of original and translated image pair.}
  \label{fig:img_translation_v1.png}
\end{figure*}

\begin{figure*}[h]
  \includegraphics[width=\textwidth]{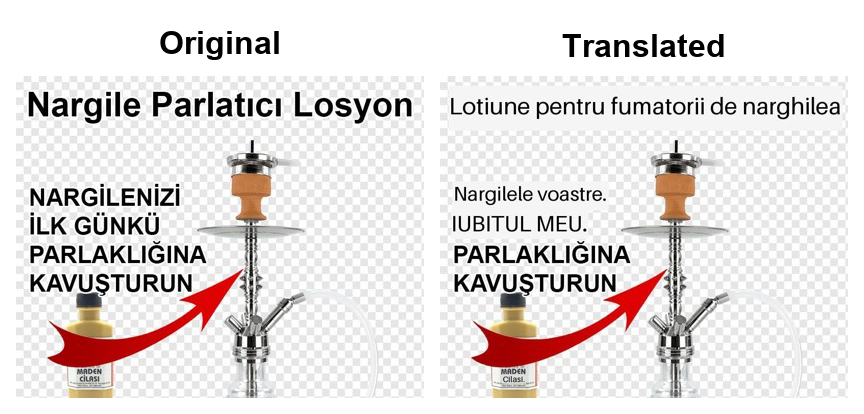}
  \caption{Example of original and translated image pair.}
  \label{fig:img_translation_v2.png}
\end{figure*}

Figure~\ref{fig:img_translation_v1.png} and Figure~\ref{fig:img_translation_v2.png} present examples of translated image pairs. It is important to note that, during the translation process, we aim to preserve the original formatting as closely as possible, including font type, font size, and text placement. Numerical values are retained without modification to prevent errors or hallucinations that could compromise critical information. While the system supports translation from multiple languages, not only English (see Figure~\ref{fig:img_translation_v2.png}), the process is not entirely foolproof, as certain portions of the text may not be correctly recognized and therefore remain untranslated.

\section{Evaluation Protocol}
\label{sec:appendix-eval-protocol}

For evaluation we integrate all tasks into lmms-eval~\cite{zhang2024lmmsevalrealitycheckevaluation}. For already existing tasks (i.e. MMBench, MMStar, MMMU, MME, SeedBench2, LLaVA-Wild) we inherit existing code with minimal changes (i.e. dataset loading, prompt language). For other benchmarks we write our own tasks definition in the lmms-eval framework. 

In the evaluation process we also integrate LLM judges where is required and we define two tasks that they solve: answer extraction and simple matching (MMBench, CoSyn, Pixmo-Count, Pixmo-Points, Flickr30k-QA) and quality judgment (LLaVA-Wild, ALM-Bench, AyaVisionBench, m-WildVision and HoraVQA). For the former we employ Qwen3-32B\footnote{https://huggingface.co/Qwen/Qwen3-32B}, while for the latter we use GPT-5.4\footnote{gpt-5.4-2026-03-05} as judge. 

Details regarding the judges used, metrics reported and prompts are presented in Table~\ref{tab:eval-protocol}.

For MME evaluation, we report the average of normalized scores for Cognition ($\text{score}/8$) and Perception ($\text{score}/20$).

\begin{table*}
  \centering
  \begin{tabular}{l c l l}
    \hline
    \textbf{Dataset} & \textbf{Metric reported} & \textbf{Judge used} & \textbf{Prompt (short)} \\
    \hline
    MMStar         & Accuracy     & -         & \multirow{6}{*}{\makecell[l]{\{Question\} \\
                                                                              \{Options\} \\
                                                                              Răspunde cu litera opțiunii alese.}} \\
    MMMU           & Accuracy     & -         & \\
    SeedBench2     & Accuracy     & -         & \\
    CVQA           & Accuracy     & -         & \\
    RoMemes        & Accuracy     & -         & \\
    MMBench        & Accuracy     & Qwen3-32B & \\
    \hline
    MME            & Accuracy     & -         & \makecell[l]{\{Question\} \\
                                                              Răspundeți cu da sau nu.} \\
    \hline
    ALM-Bench      & Score (0-10) & GPT-5.4   & \multirow{7}{*}{\{Question\}} \\
    HoraVQA      & Score (0-10) & GPT-5.4   & \\
    LLaVA-Wild     & Score (0-10) & GPT-5.4   & \\
    AyaVisionBench & Score (0-10) & GPT-5.4   & \\
    m-WildVision   & Score (0-10) & GPT-5.4   & \\
    Flickr30k-QA   & Score (0-10) & Qwen3-32B & \\
    CoSyn          & Score (0-10) & Qwen3-32B & \\
    \hline
    Flickr30k-Caption & BERTScore & -         & Descrie pe scurt această imagine. \\
    \hline
    FinePDFs       & ANSL         & -         & \multirow{2}{*}{\makecell[l]{Extrage întocmai textul și doar textul din această imagine,\\
                                                                              rezolvă task-ul de OCR.}} \\
    RoMemes-OCR    & ANSL         & -         & \\
    \hline
    Pixmo-Count    & Exact match  & Qwen3-32B & \makecell[l]{Câte \{label\} sunt în imagine? Răspundeți doar cu un număr.} \\
    \hline
    Pixmo-Points   & F1           & Qwen3-32B & \makecell[l]{Indicați toate aparițiile lui \{label\}. \\
                                                              Răspundeți cu o listă de puncte în forma ``x1'' ``y1'' \\
                                                              ``x2'' ``y2'' \dots{} sau cu «Nu există în imagine.» \\
                                                              dacă nu există niciunul.} \\
    \hline
  \end{tabular}
  \caption{\label{tab:eval-protocol}
    Evaluation benchmarks grouped by prompt template, together with the metric reported and the judge used.}
\end{table*}

\section{Annotation Process}
\label{sec:appendix-annotation-process}

Human annotation was incorporated at multiple stages throughout this work, including: (i) translation curation and (ii) benchmark construction.

Because the translation process is inherently noisy and may occasionally produce critical failures (e.g., empty outputs or repetitive text extending to the maximum token limit), we employ a semi-automatic annotation pipeline to identify and correct such issues. Specifically, translation pairs are automatically flagged when the translated text differs in length from the source text by a factor greater than 1.5×. These flagged pairs, along with a random sample of additional examples, are then reviewed by a single human annotator proficient in both English and Romanian. The annotator’s role is to verify and, where necessary, correct the translations.

HoraVQA was constructed by seven native Romanian-speaking annotators (four men and three women), who together contributed 232 unique images and 580 question--answer pairs split into 394 multiple-choice and 186 open-ended items. Annotation effort was uneven: five annotators each produced roughly 90--135 QA pairs over 24--57 images, while two contributed smaller sets (16 and 10 QA pairs), yielding a long-tailed distribution that nevertheless preserves stylistic diversity across question authors. The MCQ-to-open-ended ratio also varies by annotator, ranging from near-balanced (e.g., 61/74) to predominantly multiple-choice (e.g., 99/25), which we retain rather than rebalance in order to reflect natural authoring preferences.

\section{Training Data}
\label{sec:appendix-training-data}
Our training mixture spans five capability groups (Alignment, Captioning, General VQA \& Instruction, OCR \& Documents, Grounding) whose statistics are summarised in Table~\ref{tab:train_datasets_stats} (Qwen3-VL tokenizer). We observe substantial variation in text complexity and visual structure across domains. Mean response length (supervised tokens) ranges from 24 tokens in Alignment (very short-captions) to 367 in Captioning — a 15$\times$ gap at the group level,  Visual token usage exhibits a similarly broad range: low-resolution Alignment images consume on average 186 input tokens, whereas high-resolution document scans in OCR \& Documents consume 2{,}552, reflecting an order-of-magnitude difference in visual density under a fixed pixel budget. Interaction structure also varies sharply: Alignment, Captioning and Grounding are strictly single-turn, OCR \& Documents averages 1.46 assistant responses per sample, and General VQA \& Instruction averages 3.79. This diversity in text complexity, visual structure and turn pattern motivates evaluating Romanian vision–language models on a correspondingly broad benchmark suite, beyond the narrow captioning protocols typical of prior work.

\begin{table*}[hbt]
  \centering
  \begin{tabular}{lrrrr}
    \hline
    \textbf{Task} & \textbf{Input tokens} & \textbf{Response tokens} & \textbf{Turns} \\
    \hline
    Alignment                  &   186 &   24 & 1.00 \\
    Captioning                 & 1{,}590 &  367 & 1.00 \\
    General VQA \& Instruction &   835 &  158 & 3.79 \\
    OCR \& Documents           & 2{,}552 &  252 & 1.46 \\
    Grounding                  &   982 &   27 & 1.00 \\
    \hline
    \textbf{Total}             & 1{,}312 &  191 & 1.98 \\
    \hline
  \end{tabular}
  \caption{\label{tab:train_datasets_stats}
    Per-group statistics over the Romanian VLM training mixture (train splits only). \textit{Input tokens} = mean number of input tokens, dominated by image-token expansion and thus a proxy for visual density. \textit{Resp. tok.} = mean number of supervised label tokens per sample, a proxy for textual response complexity. \textit{Turns} = mean number of assistant responses.} 
\end{table*}

\section{Training Setup}
\label{sec:appendix-hparams}

Table~\ref{tab:hparams} summarizes the hyperparameters used across all training runs. This configuration was selected through small-scale experiments conducted on a single training dataset (llava\_mix), in which we evaluated peak learning rates of $1.0 \times 10^{-4}$, $1.0 \times 10^{-5}$, and $1.0 \times 10^{-6}$, as well as warm-up ratios of 1.0\% and 2.5\%.

One epoch for finetuning LLaVA-NeXT-Llama3-8B required 640 GPU (H200) hours, while training Qwen2-VL required 304 GPU hours. 276 hours were required for Qwen2.5-VL, with 178 required for Qwen3-VL and 170 hours for Gemma3. 

\begin{table}
  \centering
  \begin{tabular}{ll}
    \hline
    \textbf{Hyperparameter}      & \textbf{Value}                       \\
    \hline
    Vision backbone              & frozen                               \\
    Adapter                      & trainable                            \\
    Language backbone            & trainable                            \\
    \hline
    Optimizer                    & AdamW                                \\
    Weight decay                 & 0.0                                  \\
    Gradient clipping            & 1.0                                  \\
    LR schedule                  & cosine                               \\
    Peak learning rate           & $1.0\mathrm{e}{-5}$                  \\
    Minimum learning rate        & $1.0\mathrm{e}{-6}$                  \\
    Warm-up                      & 2.5\% of total steps                 \\
    Epochs                       & 1                                    \\
    \hline
    Effective batch size         & 64                                   \\
    Max sequence length          & 8192                                 \\
    \hline
    Precision                    & bf16                                 \\
    Hardware                     & $8\times$ H200                       \\
    \hline
  \end{tabular}
  \caption{\label{tab:hparams}
    Training hyperparameters, shared across all experiments.
    The vision backbone is kept frozen while the adapter and language backbone are updated.
    All runs use a single epoch with a cosine schedule and a short linear warm-up.
  }
\end{table}

\end{document}